\documentclass{article}

\usepackage{microtype}
\usepackage{graphicx}
\usepackage{subcaption}
\usepackage{booktabs} % for professional tables
\usepackage{tabularx}
\usepackage{enumitem}
\usepackage{centernot}
\usepackage{forloop}

\usepackage{booktabs}
\usepackage{tabularx}
\usepackage{multirow}
\usepackage{makecell}
\usepackage{array}
\usepackage{graphicx}

\usepackage{adjustbox}
\usepackage{xcolor}
\newcommand{\na}{--}
\newcommand{\best}[1]{\textbf{#1}}

\usepackage{hyperref}

\usepackage[preprint]{icml2026}

\usepackage{amsmath}
\usepackage{amssymb}
\usepackage{mathtools}
\usepackage{amsthm}

\usepackage[capitalize,noabbrev]{cleveref}

\theoremstyle{plain}
\newtheorem{theorem}{Theorem}[section]
\newtheorem{proposition}[theorem]{Proposition}
\newtheorem{lemma}[theorem]{Lemma}

\theoremstyle{definition}
\newtheorem{definition}[theorem]{Definition}

\theoremstyle{remark}

\usepackage{amssymb}
\usepackage{xcolor}
\usepackage{booktabs}
\usepackage[T1]{fontenc}
\usepackage[utf8]{inputenc}

\usepackage[textsize=tiny]{todonotes}

\icmltitlerunning{The Propagation Field: A Geometric Substrate Theory of Deep Learning}

\begin{document}

\twocolumn[
  \icmltitle{The Propagation Field: A Geometric Substrate Theory of Deep Learning}

  % It is OKAY to include author information, even for blind submissions: the
  % style file will automatically remove it for you unless you've provided
  % the [accepted] option to the icml2026 package.

  % List of affiliations: The first argument should be a (short) identifier you
  % will use later to specify author affiliations Academic affiliations
  % should list Department, University, City, Region, Country Industry
  % affiliations should list Company, City, Region, Country

  % You can specify symbols, otherwise they are numbered in order. Ideally, you
  % should not use this facility. Affiliations will be numbered in order of
  % appearance and this is the preferred way.
  \icmlsetsymbol{equal}{*}

  \begin{icmlauthorlist}
    \icmlauthor{Xingrui Gu}{yyy}
    % \icmlauthor{Firstname2 Lastname2}{equal,yyy,comp}
    % \icmlauthor{Firstname3 Lastname3}{comp}
    % \icmlauthor{Firstname4 Lastname4}{sch}
    % \icmlauthor{Firstname5 Lastname5}{yyy}
    % \icmlauthor{Firstname6 Lastname6}{sch,yyy,comp}
    % \icmlauthor{Firstname7 Lastname7}{comp}
    % %\icmlauthor{}{sch}
    % \icmlauthor{Firstname8 Lastname8}{sch}
    % \icmlauthor{Firstname8 Lastname8}{yyy,comp}
    %\icmlauthor{}{sch}
    %\icmlauthor{}{sch}
  \end{icmlauthorlist}

  \icmlaffiliation{yyy}{EECS, University of California, Berkeley, Berkeley, CA}
  % \icmlaffiliation{comp}{Company Name, Location, Country}
  % \icmlaffiliation{sch}{School of ZZZ, Institute of WWW, Location, Country}

  \icmlcorrespondingauthor{Xingrui Gu}{xingrui\_gu@berkeley.edu}
  % \icmlcorrespondingauthor{Firstname2 Lastname2}{first2.last2@www.uk}

  % You may provide any keywords that you find helpful for describing your
  % paper; these are used to populate the "keywords" metadata in the PDF but
  % will not be shown in the document
  \icmlkeywords{Machine Learning, ICML}

  \vskip 0.3in
]

% this must go after the closing bracket ] following \twocolumn[ ...

% This command actually creates the footnote in the first column listing the
% affiliations and the copyright notice. The command takes one argument, which
% is text to display at the start of the footnote. The \icmlEqualContribution
% command is standard text for equal contribution. Remove it (just {}) if you
% do not need this facility.

% Use ONE of the following lines. DO NOT remove the command.
% If you have no special notice, KEEP empty braces:
\printAffiliationsAndNotice{}  % no special notice (required even if empty)
% Or, if applicable, use the standard equal contribution text:
% \printAffiliationsAndNotice{\icmlEqualContribution}

\begin{abstract}
Modern deep learning treats neural networks primarily as endpoint
functions from inputs to outputs. Inspired by the shift from force to
geometry in physics, we ask whether a network should instead be
understood through the geometry of its internal propagation. We define a neural propagation field as the collection of hidden-state
trajectories and local Jacobian operators across depth. Endpoint losses constrain only the boundary behavior of this field, leaving its
interior geometry underdetermined. We show that endpoint-equivalent models can differ by orders of magnitude in trajectory and Jacobian
structure, and introduce observable field metrics such as path
sensitivity, solver consistency, and trajectory/Jacobian retention. In
controlled teacher-flow and PDE systems, endpoint fitting fails to
recover the underlying propagation law. In real multi-path tasks,
field-aware objectives improve unseen-path generalization, OOD
robustness, and calibration when aligned with the observation structure,
but can collapse when over-constrained. In continual learning,
field-preservation regularization complements replay and distillation:
on Split CIFAR-100, DER++ with field preservation improves average
accuracy, backward transfer, and field-retention metrics. These results
identify propagation-field quality as a measurable and trainable
property of neural networks beyond endpoint performance.
\end{abstract}

\section{Introduction}

Deep learning succeeds by optimizing input--output maps through task loss, yet how information propagates internally remains poorly understood \citep{lecun2015deep}. Standard training treats hidden layers as intermediate computations driven by output error \citep{rumelhart1986learning}, constraining endpoint behavior but not the geometry of layerwise representations. This gap appears across adversarial robustness \citep{szegedy2013intriguing,goodfellow2014explaining}, in-context learning \citep{brown2020language,olsson2022context}, continual forgetting \citep{kirkpatrick2017overcoming}, and endpoint-loss scaling laws \citep{kaplan2020scaling,hoffmann2022training}, suggesting that model behavior depends on internal propagation structure as well as endpoint functions.

Inspired by classical field theory, general relativity, and Yang--Mills gauge theory, we propose a propagation-field perspective for understanding the internal mechanisms of deep learning. In physics, a central conceptual transition from Newtonian mechanics to general relativity and non-Abelian gauge theory is to reinterpret apparent ``forces'' as manifestations of more fundamental fields and geometric structures: gravity can be described as spacetime curvature, while fundamental interactions can be formulated through the curvature of gauge fields \citep{albert1916foundation,PhysRev.96.191,nakahara2018geometry}. We view layerwise Jacobians, hidden-state trajectories, and their geometric properties path sensitivity, curvature, and velocity alignment as defining an internal propagation field that underlies the endpoint function. The endpoint function is then a terminal projection of this field, not the entirety of what learning produces. 

% \begin{enumerate}
%     \item We define measurable propagation-field quantities---hidden trajectories, local Jacobians, and their layerwise smoothness---and show that these are distinct from endpoint behavior.

%     \item We show that field-aware objectives improve generalization on path-dependent tasks but can over-regularize on endpoint-like tasks, establishing when propagation-field structure is beneficial.

%     \item We show that field-preservation regularization complements replay-based continual learning, simultaneously improving accuracy, backward transfer, and Jacobian retention on Split CIFAR-100.
% \end{enumerate}

\section{Endpoint Supervision Does Not Identify the Propagation Field}
\label{sec:underdetermination}

Standard supervised learning optimizes a task loss over a data distribution $\mathcal{D}$ \citep{lecun2015deep}:
\begin{equation}
\mathcal{L}_{\mathrm{task}}(\theta)
=
\mathbb{E}_{(x,y)\sim\mathcal{D}}
\left[
\ell(f_\theta(x),y)
\right].
\end{equation}
This constrains input--output behavior but does not uniquely determine internal propagation.

\begin{definition}[Endpoint Equivalence]
Given a task distribution $\mathcal{D}$ and tolerance $\epsilon>0$, two models $f_{\theta_1}$ and $f_{\theta_2}$ are \emph{endpoint-equivalent} if
\begin{equation}
\left|
\mathcal{L}_{\mathrm{task}}(\theta_1)
-
\mathcal{L}_{\mathrm{task}}(\theta_2)
\right|
\leq \epsilon .
\end{equation}
\end{definition}

Endpoint-equivalent models need not propagate information the same way. We denote the propagation field of a deep model by
\begin{equation}
\Phi_\theta(x)
=
\{h_0,h_1,\ldots,h_L;\; J_0,J_1,\ldots,J_{L-1}\},
\end{equation}
where $h_\ell$ is the hidden state at layer $\ell$ and
\begin{equation}
J_\ell
=
\frac{\partial h_{\ell+1}}{\partial h_\ell}
\end{equation}
is the local propagation Jacobian. For continuous-flow models, $\Phi_\theta$ may also include the trajectory $h(t)$, the vector field $f_\theta(h,t)$, and its numerical integration behavior.

\begin{definition}[Propagation-Field Distance]
Let $\Phi_{\theta_1}$ and $\Phi_{\theta_2}$ be the propagation fields of two models $f_{\theta_1}$ and $f_{\theta_2}$. A field distance is
\begin{equation}
\label{eq:field_dist}
\begin{aligned}
d_{\text{field}}(\theta_1, \theta_2) = \mathbb{E}_{x \sim \mathcal{D}} \big[ &d_{\text{traj}} (\mathcal{H}_{\theta_1}(x), \mathcal{H}_{\theta_2}(x)) \\
&+ \lambda_J d_{\text{Jac}} (\mathcal{J}_{\theta_1}(x), \mathcal{J}_{\theta_2}(x)) \big]
\end{aligned}
\end{equation}
where $\mathcal{H}_\theta(x)$ denotes the hidden-state trajectory and $\mathcal{J}_\theta(x)$ denotes the collection of local Jacobians or their spectral proxies.
\end{definition}

\begin{proposition}[Endpoint Equivalence Does Not Imply Field Equivalence]
A small discrepancy in task loss does not imply a small discrepancy in the propagation field:
\begin{equation}
\mathcal{L}_{\mathrm{task}}(\theta_1)
\approx
\mathcal{L}_{\mathrm{task}}(\theta_2)
\centernot\Rightarrow
d_{\mathrm{field}}(\theta_1,\theta_2)
\approx 0.
\end{equation}
\end{proposition}

Scaling laws therefore characterize the scaling of endpoint loss, but not necessarily the scaling of internal propagation quality \citep{kaplan2020scaling,hoffmann2022training}.

\subsection{Experimental Evidence}
Since standard benchmarks rarely measure internal propagation, we provide compact empirical evidence in Table~\ref{tab:endpoint_nonidentifiability}. Experiment details are in Appendices~\ref{app:teacher_flow}, \ref{app:pde_metrics}, and \ref{app:reveal_path_transfer}. The teacher-flow setting gives the cleanest test: models with nearly identical endpoint accuracy can exhibit orders-of-magnitude differences in trajectory and derivative recovery. PDE extrapolation shows that endpoint fitting does not recover propagation laws, while reveal-path tasks indicate that path sensitivity is a distinct internal axis rather than a proxy for accuracy. These results support the claim that endpoint supervision identifies a behavioral equivalence class but does not uniquely determine the internal propagation field.

\begin{table}[t]
\centering
\caption{Endpoint equivalence does not imply field equivalence. Endpoint metrics remain similar, while field metrics can differ substantially. Lower field values are better.}
\label{tab:endpoint_nonidentifiability}
\scriptsize
\setlength{\tabcolsep}{3.5pt}
\begin{tabular}{lccc}
\toprule
\textbf{Setting} & \textbf{Endpoint} & \textbf{Field} & \textbf{Gap} \\
\midrule
Teacher A, Traj.
& \(0.9975/0.9975\)
& \(3.52 \!\to\! 0.06\)
& \(58.7\times\) \\
Teacher A, Deriv.
& \(0.9975/0.9975\)
& \(2.76 \!\to\! 0.15\)
& \(18.4\times\) \\
Teacher B, Traj.
& \(\approx .99/\approx .99\)
& \(3.33 \!\to\! 0.07\)
& \(48.0\times\) \\
PDE-A, \(T{=}2.0\)
& endpoint fit
& \(1.217 \!\to\! 0.195\)
& \(6.2\times\) \\
Tiny-ImageNet
& \(.097 \!\to\! .118\)
& \(4.49 \!\to\! 1.27\)
& \(3.5\times\) \\
\bottomrule
\end{tabular}
\vspace{-0.5em}
\end{table}

\section{Propagation-Field Theory}
\label{sec:field_theory}

The previous section showed that endpoint-equivalent models can have different hidden trajectories, Jacobian structures, and path sensitivities. We now define the measurable quantities that characterize these differences. Rather than assuming Riemannian structure or geodesic dynamics, we work directly with quantities computable in standard networks: hidden-state trajectories and local Jacobian operators.

\begin{definition}[Observable Propagation Path]
Given an input $x$ and model $f_\theta$, let
\begin{equation}
h_0^\theta(x), h_1^\theta(x), \ldots, h_L^\theta(x)
\end{equation}
be the hidden states across depth, where $h_0^\theta(x)$ is the input encoding and $h_L^\theta(x)$ is the representation before the prediction head. Since hidden dimensions may vary across layers, we introduce a layer embedding map $\phi_\ell:\mathcal H_\ell \rightarrow \mathbb R^d$ and define the standardized hidden state
\begin{equation}
z_\ell^\theta(x)=\phi_\ell(h_\ell^\theta(x)).
\end{equation}
If all layers share the same dimension, $\phi_\ell$ is the identity. The \textbf{observable propagation path} is
\begin{equation}
\gamma_\theta(x)
=
\{z_0^\theta(x),z_1^\theta(x),\ldots,z_L^\theta(x)\}.
\end{equation}
\end{definition}

\begin{definition}[Local Propagation Operator]
At layer $\ell$, the local propagation operator is
\begin{equation}
J_\ell^\theta(x)
=
\frac{\partial h_{\ell+1}^\theta(x)}
{\partial h_\ell^\theta(x)}.
\end{equation}
For a perturbation $v$ at layer $\ell$, the first-order change at layer $\ell+1$ is $\delta h_{\ell+1} \approx J_\ell^\theta(x)v$. Thus $J_\ell^\theta(x)$ describes how a single layer amplifies, compresses, or redirects perturbations in hidden space.
\end{definition}

\begin{definition}[Jacobian-Induced Propagation Metric]
The local Jacobian induces a layerwise metric
\begin{equation}
G_\ell^\theta(x)
=
(J_\ell^\theta(x))^\top J_\ell^\theta(x),
\end{equation}
for which $v^\top G_\ell^\theta(x)v = \|J_\ell^\theta(x)v\|_2^2$. Strong amplification indicates sensitivity; strong compression indicates information loss. In high-dimensional networks, $G_\ell^\theta(x)$ is not formed explicitly; experiments use scalable proxies such as Jacobian-vector products, random projections, and spectral estimates.
\end{definition}

\subsection{Accumulated Propagation Metric and Global Perturbation Transport}

\begin{definition}[Accumulated Propagation Operator]
The accumulated Jacobian from layer $0$ to layer $\ell$ is
\begin{equation}
\bar J_{0:\ell}^\theta(x)
=
J_{\ell-1}^\theta(x)
\cdots
J_1^\theta(x)
J_0^\theta(x),
\end{equation}
with the corresponding accumulated metric $\bar G_{0:\ell}^\theta(x) = (\bar J_{0:\ell}^\theta(x))^\top \bar J_{0:\ell}^\theta(x)$.
\end{definition}

\begin{lemma}[First-Order Perturbation Propagation]
For a small input perturbation $\delta h_0$, the displacement at depth $\ell$ satisfies
\begin{equation}
h_\ell^\theta(x+\delta x)-h_\ell^\theta(x)
\approx
\bar J_{0:\ell}^\theta(x)\,\delta h_0,
\end{equation}
and therefore
\begin{equation}
\|h_\ell^\theta(x+\delta x)-h_\ell^\theta(x)\|_2^2
\approx
\delta h_0^\top
\bar G_{0:\ell}^\theta(x)
\delta h_0.
\end{equation}
The spectral structure of $\bar G_{0:\ell}^\theta(x)$ therefore governs perturbation sensitivity: a large eigenvalue along a given direction means small input perturbations in that direction can be amplified through depth. This is a first-order approximation; for large perturbations it serves as a geometric explanation rather than an exact identity.
\end{lemma}

% \subsection{Observable Geometry of Propagation Paths}

To make propagation fields measurable, we use three discrete geometry metrics.

\paragraph{Path length.}
\begin{equation}
\mathrm{Len}_\theta(x)
=
\sum_{\ell=0}^{L-1}
\left\|
z_{\ell+1}^{\theta}(x)-z_{\ell}^{\theta}(x)
\right\|_2 .
\end{equation}
This measures the total representation displacement across depth.

\paragraph{Discrete curvature.}
Let
\begin{equation}
\Delta z_\ell^\theta(x)
=
z_{\ell+1}^{\theta}(x)-z_{\ell}^{\theta}(x).
\end{equation}
We define
\begin{equation}
\kappa_\theta(x)
=
\frac{1}{L-1}
\sum_{\ell=1}^{L-1}
\frac{
\left\|
z_{\ell+1}^{\theta}(x)
-
2z_{\ell}^{\theta}(x)
+
z_{\ell-1}^{\theta}(x)
\right\|_2
}{
\left\|
z_{\ell+1}^{\theta}(x)-z_{\ell}^{\theta}(x)
\right\|_2^2
+
\varepsilon
}.
\end{equation}
This is a discrete proxy for the bending of the hidden trajectory, not a strict Riemannian curvature.

\paragraph{Velocity alignment.}
\begin{equation}
\mathrm{Align}_\theta(x)
=
\frac{1}{L-1}
\sum_{\ell=1}^{L-1}
\frac{
\left\langle
\Delta z_\ell^\theta(x),
\Delta z_{\ell-1}^\theta(x)
\right\rangle
}{
\left\|
\Delta z_\ell^\theta(x)
\right\|_2
\left\|
\Delta z_{\ell-1}^\theta(x)
\right\|_2
+
\varepsilon
}.
\end{equation}
High alignment indicates consistent propagation direction; low or negative alignment indicates oscillation or reversal. These metrics correspond to Path, Curvature, and Velocity Align in our experiments.

% \subsection{Path Sensitivity under Multi-path Inputs}

Many tasks admit multiple equivalent observation paths, such as image patch orders, frequency decompositions, temporal speech windows, or different textual evidence orders. For two paths \(p,q\) of the same sample \(x\), let
\begin{equation}
h_L^{\theta,p}(x),
\quad
h_L^{\theta,q}(x),
\quad
o^{\theta,p}(x),
\quad
o^{\theta,q}(x)
\end{equation}
denote the final hidden states and logits.

We define hidden-state path sensitivity as
\begin{equation}
\mathrm{PathSens}_h
=
\mathbb E_{x,p,q}
\left[
\left\|
h_L^{\theta,p}(x)
-
h_L^{\theta,q}(x)
\right\|_2
\right],
\end{equation}
and output-level path sensitivity as
\begin{equation}
\mathrm{PathSens}_o
=
\mathbb E_{x,p,q}
\left[
\left\|
o^{\theta,p}(x)
-
o^{\theta,q}(x)
\right\|_2
\right].
\end{equation}
Lower path sensitivity indicates greater invariance to observation paths, but it does not guarantee better task performance: collapsed representations can also yield low sensitivity. We therefore evaluate it together with accuracy, OOD generalization, calibration, and representation variance.

% \subsection{Solver Consistency and Layer Reparameterization}

For shared-field or continuous-depth models, the same underlying field can be evaluated under different discretizations. Let
\begin{equation}
h_T^{(s)}(x),
\quad
o_T^{(s)}(x)
\end{equation}
be the final hidden state and logits under solver or step configuration \(s\). We define
\begin{equation}
\begin{aligned}
&\mathrm{SolverErr}
=\\
&\mathbb E_{x,s_1,s_2}
\left[
\left\|
h_T^{(s_1)}(x)
-
h_T^{(s_2)}(x)
\right\|_2 
+
\left\|
o_T^{(s_1)}(x)
-
o_T^{(s_2)}(x)
\right\|_2
\right].
\end{aligned}
\end{equation}
Low solver error, especially under step refinement, suggests that layers behave as numerical slices of a shared propagation field rather than as a fixed-depth black box.

% \subsection{Propagation Field Retention in Continual Learning}

In continual learning, new tasks may overwrite not only the output function but also the internal propagation field of old tasks. Let \(\theta^{\mathrm{old}}\) and \(\theta^{\mathrm{new}}\) denote parameters before and after learning a new task, and let \(\mathcal A\) be old-task anchor samples.

\begin{definition}[Trajectory Retention Score]
\begin{equation}
\mathrm{FRS}
=
\mathbb E_{x\in\mathcal A}
\frac{1}{L+1}
\sum_{\ell=0}^{L}
\left\|
h_\ell^{\theta^{\mathrm{old}}}(x)
-
h_\ell^{\theta^{\mathrm{new}}}(x)
\right\|_2^2.
\end{equation}
Lower FRS indicates better preservation of old-task hidden trajectories.
\end{definition}

\begin{definition}[Jacobian Retention Score]
\begin{equation}
\mathrm{JRS}
=
\mathbb E_{x\in\mathcal A,\ell,\delta}
\left[
\frac{
\left\|
J_\ell^{\theta^{\mathrm{old}}}(x)\delta
-
J_\ell^{\theta^{\mathrm{new}}}(x)\delta
\right\|_2^2
}{
\|\delta\|_2^2+\varepsilon
}
\right].
\end{equation}
Lower JRS indicates better preservation of old-task local propagation operators.
\end{definition}

FRS measures hidden-trajectory drift, while JRS measures Jacobian-field drift. They capture field-level forgetting, complementing standard function-level continual learning metrics such as AA and BWT.

\subsection{Field-aware Optimization Objectives}

Propagation geometry can also be optimized. We define multi-path consistency as
\begin{equation}
\label{eq:reveal_loss}
\begin{aligned}
\mathcal{L}_{\text{reveal}} = \mathbb{E}_{x, p, q} \Big[ & \left\| h_L^{\theta, p}(x) - h_L^{\theta, q}(x) \right\|_2^2 \\
&+ \text{KL} \big( \sigma(o^{\theta, p}(x)) \,\|\, \sigma(o^{\theta, q}(x)) \big) \Big].
\end{aligned}
\end{equation}
Solver consistency is
\begin{equation}
\label{eq:solver_loss}
\begin{aligned}
\mathcal{L}_{\text{solver}} = \mathbb{E}_{x, s_1, s_2} \Big[ &\left\| h_T^{(s_1)}(x) - h_T^{(s_2)}(x) \right\|_2^2 \\
+ &\left\| o_T^{(s_1)}(x) - o_T^{(s_2)}(x) \right\|_2^2 \Big].
\end{aligned}
\end{equation}
Jacobian-field smoothness is
\begin{equation}
\mathcal L_{\mathrm{jac}}
=
\mathbb E_{x,\ell,\delta}
\left[
\frac{
\left\|
J_\ell^\theta(x)\delta
-
J_{\ell-1}^\theta(x)\delta
\right\|_2^2
}{
\|\delta\|_2^2+\varepsilon
}
\right].
\end{equation}
The general field-aware loss is
\begin{equation}
\mathcal L_{\mathrm{field}}
=
\lambda_r \mathcal L_{\mathrm{reveal}}
+
\lambda_s \mathcal L_{\mathrm{solver}}
+
\lambda_J \mathcal L_{\mathrm{jac}},
\end{equation}
and the training objective is
\begin{equation}
\mathcal L_{\mathrm{total}}
=
\mathcal L_{\mathrm{task}}
+
\mathcal L_{\mathrm{field}}.
\end{equation}

For continual learning, we combine field preservation with replay or distillation:
\begin{equation}
\mathcal L_{\mathrm{CL}}
=
\mathcal L_{\mathrm{new}}
+
\mathcal L_{\mathrm{replay/distill}}
+
\lambda_{\mathrm{FPR}}
\mathcal L_{\mathrm{FPR}},
\end{equation}
where
\begin{equation}
\begin{aligned}
\mathcal L_{\mathrm{FPR}}
=
&\lambda_h
\mathbb E_{x,\ell}
\left[
\left\|
h_\ell^{\theta^{\mathrm{old}}}(x)
-
h_\ell^\theta(x)
\right\|_2^2
\right]
\\
&+
\lambda_J
\mathbb E_{x,\ell,\delta}
\left[
\frac{
\left\|
J_\ell^{\theta^{\mathrm{old}}}(x)\delta
-
J_\ell^\theta(x)\delta
\right\|_2^2
}{
\|\delta\|_2^2+\varepsilon
}
\right].
\end{aligned}
\end{equation}
FPR is not a replacement for replay or distillation; it adds the missing propagation-structure constraint. Conventional methods preserve old-task discriminative functions, while FPR preserves their internal propagation geometry.

\subsection{Field Collapse and Discriminative Constraints}

Field-aware regularization can be harmful when it is too strong. A model exhibits field collapse when field-consistency metrics improve while discriminative performance degrades. Typical indicators include
\begin{equation}
\mathrm{Var}_{x}
\left[
h_L^\theta(x)
\right]
\rightarrow 0,
\end{equation}
or prediction homogenization:
\begin{equation}
p_\theta(y\mid x)
\approx
p_0(y),
\quad
\forall x.
\end{equation}
Thus,
\begin{equation}
\mathrm{PathSens}\downarrow
\centernot\Rightarrow
\mathrm{Accuracy}\uparrow.
\end{equation}
Effective field-aware learning must therefore combine propagation consistency with discriminative and anti-collapse constraints.

\subsection{Experiment}

\begin{table*}[t]
\centering
\caption{
\textbf{Compact experimental evidence for propagation-field geometry.}
We report representative results across controlled propagation, real multi-path transfer, collapse cases, and continual learning. 
Task metrics measure endpoint behavior or function retention; field metrics measure hidden trajectory, local Jacobian, or path consistency. Lower field metrics are better.
}
\label{tab:main_propagation_evidence}
\scriptsize
\setlength{\tabcolsep}{3.2pt}
\renewcommand{\arraystretch}{1.08}
\begin{tabular}{p{2.5cm}p{2.25cm}p{2.55cm}p{3.0cm}p{3.25cm}p{3.0cm}}
\toprule
\textbf{Claim} 
& \textbf{Setting} 
& \textbf{Comparison} 
& \textbf{Task / Endpoint Metric} 
& \textbf{Generalization / Calibration} 
& \textbf{Propagation-Field Metric} \\
\midrule

Endpoint does not identify field
& Teacher-flow A
& FlowEndOnly $\rightarrow$ FlowFieldLoss
& Acc $.9975 \rightarrow .9975$
& Reparam. final $2.45{\times}10^{-3} \rightarrow 1.23{\times}10^{-3}$
& Traj. $3.52 \rightarrow .059$; Deriv. $2.76 \rightarrow .150$ \\

Endpoint fitting does not recover law
& PDE-A extrapolation
& Endpoint map $\rightarrow$ generator-style field
& Trained endpoint fit
& $T{=}2$ MSE $1.217 \rightarrow .195$
& Energy corr. N/A $\rightarrow .995$; regrid err. $2.17{\times}10^{-2}$ \\

Field helps when task-aligned
& Tiny-ImageNet reveal
& Task $\rightarrow$ Full field-aware
& Acc $.097 \rightarrow .118$
& Unseen $.085 \rightarrow .109$; OOD $.100 \rightarrow .128$; ECE $.102 \rightarrow .018$
& PathSens-logit $4.49 \rightarrow 1.27$; hidden $4.44 \rightarrow 2.07$ \\

Over-constraint can collapse
& CIFAR-100 reveal
& Task $\rightarrow$ Full field-aware
& Acc $.0405 \rightarrow .0093$
& Unseen $.0295 \rightarrow .0090$; OOD $.057 \rightarrow .010$
& PathSens-logit $15.19 \rightarrow 3.04$; hidden $7.28 \rightarrow 2.14$ \\

Field retention differs from function retention
& Split CIFAR-100 CL
& DER++ vs. FPR-Traj
& AA $.1346$ vs. $.0374$
& BWT strong for replay; weak for standalone FPR
& FRS $.9209 \rightarrow .5568$; JRS $.1768 \rightarrow .0332$ \\

FPR complements replay/distillation
& Split CIFAR-100 CL, budget 200
& DER++ $\rightarrow$ DER++ + FPR-Full
& AA $.125 \rightarrow .136$; BWT $-.538 \rightarrow -.481$
& Hybrid improves function retention
& FRS $.980 \rightarrow .714$; JRS $.166 \rightarrow .091$ \\

\bottomrule
\end{tabular}
\vspace{-0.5em}
\end{table*}

We evaluate both endpoint behavior and internal propagation geometry, organized around four questions: \textbf{Q1} whether endpoint supervision uniquely identifies the propagation field; \textbf{Q2} whether field-aware objectives improve multi-path, OOD, or calibration performance; \textbf{Q3} whether field consistency can over-regularize and cause collapse; and \textbf{Q4} whether field preservation improves continual learning when combined with replay or distillation. Experiment details are in Appendix~\ref{app:experimental_overview}.

\paragraph{Main Analysis }Table~\ref{tab:main_propagation_evidence} summarizes six representative results, ordered from controlled systems to real tasks and continual learning. In teacher-flow, two models achieve identical accuracy \((0.9975)\) but differ by \(59\times\) in trajectory recovery: field supervision reduces trajectory error from \(3.52\) to \(0.06\) without changing endpoint performance. In PDE extrapolation, endpoint maps fit the training horizon but fail at \(T=2.0\) \((\mathrm{MSE}=1.217)\), whereas generator-style propagation models reach \(\mathrm{MSE}=0.195\) with energy correlation \(0.995\). These results show that endpoint loss identifies a behavioral equivalence class, not the underlying propagation law. On real multi-path tasks, field-aware objectives help only when aligned with task structure. On Tiny-ImageNet, the full objective improves accuracy \((0.097\to0.118)\), unseen-path accuracy, OOD accuracy, and calibration while reducing path sensitivity. On CIFAR-100, however, it reduces path sensitivity but collapses accuracy to \(0.0093\), showing that field consistency alone is not sufficient. Continual learning shows the same complementarity: standalone FPR preserves field metrics \((\mathrm{FRS}=0.557,\mathrm{JRS}=0.033)\) but not endpoint performance \((\mathrm{AA}=0.037)\), while DER++ preserves endpoint performance \((\mathrm{AA}=0.135)\) but not field geometry \((\mathrm{JRS}=0.177)\). Combining them recovers both: DER++ + FPR-Full at \(200\) samples/task improves AA \((0.125\to0.136)\), BWT \((-0.538\to-0.481)\), and JRS \((0.166\to0.091)\), indicating that field retention and function retention are complementary objectives.

\paragraph{Controlled Evidence: Endpoint Behavior Does Not Identify the Field}

Table~\ref{tab:app_pde_summary} summarizes the controlled evidence. In teacher-flow systems, endpoint accuracy remains essentially unchanged after adding field supervision, yet trajectory and derivative errors drop by one to two orders of magnitude. This directly supports the underdetermination claim: endpoint-equivalent models need not be field-equivalent. In PDE systems, endpoint maps achieve very low training-horizon error but fail at \(T=2.0\), whereas propagation-style models substantially reduce extrapolation error and preserve energy evolution. Thus, fitting endpoints is not equivalent to learning the underlying propagation law.

\paragraph{Real Multi-path Transfer: When Field Objectives Help}
We test whether field-aware objectives improve performance on real multi-path tasks. Table~\ref{tab:app_real_transfer} shows that gains are task-dependent. Tiny-ImageNet provides the clearest positive result: the full objective improves AccAvg, UnseenAcc, OOD accuracy, and ECE while reducing both logit and hidden path sensitivity. CIFAR-100 shows a weaker signal: reveal consistency improves accuracy and lowers PathSens, but the full objective degrades accuracy despite further reducing path sensitivity. AG News and SST-2 show no consistent improvement, with the full objective reducing AG News accuracy to near-random. Speech Commands yields modest OOD gains but limited ID improvement. Overall, field-aware objectives help when path consistency aligns with task structure, but can collapse discriminative performance when imposed without task-aligned constraints.

\paragraph{Field Collapse: Low Path Sensitivity Is Not Sufficient} As shown in Table~\ref{tab:app_real_transfer}, lower path sensitivity does not necessarily imply better task performance. For example, the full objective on CIFAR-100 reduces PathSensLogit from \(15.1867\) to \(3.0355\), but AccAvg drops from \(0.0405\) to \(0.0093\). Similarly, AG News already has low PathSens under task-only training, while reveal/full objectives degrade accuracy.

\paragraph{Field-aware Objectives and Optimization Tradeoffs}
Table~\ref{tab:objective_ablations} summarizes objective-level effects, and Table~\ref{tab:app_strict_suite_summary} summarizes training-level Pareto behavior. Reveal consistency is the strongest single component, increasing AccAvg from $0.3472$ to $0.8461$ and reducing PathSensLogit from $25.6669$ to $6.9140$. The full objective yields the lowest PathSensHidden ($3.9182$) and JacWDist ($0.0032$), but does not exceed reveal-only accuracy. At the optimization level, different algorithms occupy different Pareto points: LocalFieldMatch improves over FullBPTT in both TestAcc ($0.615$ vs.\ $0.590$) and TrajRMSE ($1.034$ vs.\ $1.053$), while gradient-cosine analysis shows task--field conflict with negative-gradient fractions up to $0.36$. Standard end-to-end backpropagation does not automatically find the best task--field trade-off.

\paragraph{Continual Learning: Field Retention as a Complementary Objective}
Table~\ref{tab:app_cl_fpr_all} shows that field retention complements function retention in Split CIFAR-100. Trajectory drift is weakly but significantly correlated with forgetting ($r=0.1748$, $p=0.0158$; $\rho=0.1498$, $p=0.0391$), while Jacobian and parameter drift are not significant predictors. Replay and distillation methods are stronger for endpoint retention: DER++ achieves AA $0.1346$ and BWT $-0.5248$. Standalone FPR mainly improves field metrics---FPR-Traj obtains the lowest JRS ($0.0332$). The strongest result is the hybrid setting: DER++ + FPR-Full at 200 samples/task improves AA, BWT, FRS, and JRS simultaneously ($+0.0111$, $+0.0571$, $-0.2661$, $-0.0748$). FPR is not a replacement for replay, but a field-retention regularizer that complements existing methods.

\section{Conclusion}

We argue that a deep network is an endpoint function at its boundary, but a propagation field in its interior. We formalize this field through hidden trajectories and local Jacobian operators, show that endpoint supervision underdetermines it, and introduce metrics and objectives to measure, regularize, and preserve it. Across controlled, multi-path, and continual learning settings, propagation-field quality emerges as a distinct, trainable property beyond endpoint performance. Finally, our collapse results suggest a Kakeya-like directional principle: useful propagation fields must reduce spurious path sensitivity while preserving sufficient directional richness for discrimination \citep{wang2026sticky}.

\newpage
\newpage
\bibliography{example_paper}
\bibliographystyle{icml2026}

%%%%%%%%%%%%%%%%%%%%%%%%%%%%%%%%%%%%%%%%%%%%%%%%%%%%%%%%%%%%%%%%%%%%%%%%%%%%%%%
%%%%%%%%%%%%%%%%%%%%%%%%%%%%%%%%%%%%%%%%%%%%%%%%%%%%%%%%%%%%%%%%%%%%%%%%%%%%%%%
% APPENDIX
%%%%%%%%%%%%%%%%%%%%%%%%%%%%%%%%%%%%%%%%%%%%%%%%%%%%%%%%%%%%%%%%%%%%%%%%%%%%%%%
%%%%%%%%%%%%%%%%%%%%%%%%%%%%%%%%%%%%%%%%%%%%%%%%%%%%%%%%%%%%%%%%%%%%%%%%%%%%%%%
\newpage
\appendix
\onecolumn
\section{Experimental Overview and Unified Protocol}
\label{app:experimental_overview}

This series of experiments is designed around a central question: whether the internal computation of deep learning models can be understood as a \emph{propagation field} evolving along depth, time, or observation paths, and whether such a field can serve as a measurable, trainable, and retainable object in machine learning. Unlike the conventional endpoint-function view, which primarily evaluates the mapping from inputs to outputs, our experimental framework explicitly examines hidden trajectories, local Jacobians, path consistency, solver consistency, and cross-task retention of internal propagation structures.

\subsection{Core Hypotheses}
\label{app:core_hypotheses}

We organize the experiments around the following hypotheses:

\begin{itemize}[leftmargin=1.5em]
    \item \textbf{Endpoint supervision is underdetermined.} 
    Standard task losses typically constrain only final labels or endpoint outputs. As a result, they may identify a behavioral equivalence class rather than a unique internal propagation field.

    \item \textbf{Propagation quality can become useful when aligned with task structure.} 
    When a task involves path dependence, sequential dependence, partial observability, multi-view evidence, or long-horizon process structure, the quality of the internal propagation field may translate into measurable learning benefits.

    \item \textbf{Propagation quality is multi-dimensional.} 
    It cannot be reduced to a single scalar measure, but is jointly characterized by state trajectories, local propagation operators, path consistency, and numerical reparameterization consistency.

    \item \textbf{Forgetting may occur at the field level.} 
    In continual learning, forgetting may not only appear as degradation of endpoint performance, but also as rewriting of the internal propagation field associated with previous tasks. However, field retention and accuracy retention are not necessarily equivalent.
\end{itemize}

\subsubsection{Unified Experimental Objects}
\label{app:experimental_objects}

Across experiments, we compare several model families that instantiate different assumptions about endpoint mappings and internal propagation structures. Table~\ref{tab:experimental_objects} summarizes their roles.

\begin{table}[t]
\centering
\caption{Unified experimental objects used across the study.}
\label{tab:experimental_objects}
\begin{tabularx}{\linewidth}{p{0.25\linewidth} p{0.38\linewidth} X}
\toprule
\textbf{Object} & \textbf{Definition} & \textbf{Role in Experiments} \\
\midrule
Endpoint Map 
& Directly learns a mapping from inputs to labels or endpoint states. 
& Serves as the baseline corresponding to the endpoint-function view. \\
\midrule
Time-Conditioned Discrete Flow 
& Learns a family of discrete propagation maps conditioned on time or layer index. 
& Tests whether adding temporal or depth conditioning improves internal propagation structure. \\
\midrule
Shared-Field Discrete Flow 
& Shares the same propagation parameters across layers and advances states through Euler-like updates. 
& Tests whether layers can be interpreted as discrete slices of a common propagation field. \\
\midrule
Continuous Flow 
& Explicitly learns a continuous vector field and obtains terminal states through numerical integration. 
& Evaluates solver and step-size reparameterization consistency. \\
\midrule
Residual Stack 
& Uses a standard residual network with independent parameters at each layer. 
& Serves as a conventional deep discrete-architecture baseline. \\
\midrule
Hybrid Flow 
& Combines a shared propagation field with a small number of learnable correction terms. 
& Balances continuous-field constraints with expressive flexibility. \\
\bottomrule
\end{tabularx}
\end{table}

\subsubsection{Unified Evaluation Metrics}
\label{app:evaluation_metrics}

We evaluate models along two complementary axes: external task performance and internal field quality. The former measures whether the model produces correct endpoint behavior, while the latter measures how information is propagated internally. Table~\ref{tab:evaluation_metrics} summarizes the metrics used throughout the experiments.

\begin{table}[t]
\centering
\caption{Unified evaluation metrics for endpoint performance and internal propagation quality.}
\label{tab:evaluation_metrics}
\begin{tabularx}{\linewidth}{p{0.24\linewidth} p{0.28\linewidth} X}
\toprule
\textbf{Category} & \textbf{Metric} & \textbf{Meaning} \\
\midrule
External Task Metrics 
& ID Accuracy 
& Classification or prediction accuracy on the in-distribution test set. \\
\midrule
External Task Metrics 
& OOD Accuracy 
& Accuracy under distributional perturbations, such as noise, path shifts, center shifts, rotations, or truncations. \\
\midrule
External Task Metrics 
& Unseen-Path Accuracy 
& Accuracy on reveal paths or observation paths not encountered during training. \\
\midrule
External Task Metrics 
& Low-Shot Accuracy 
& Task performance under few-shot training conditions. \\
\midrule
External Task Metrics 
& ECE 
& Expected Calibration Error, used to evaluate confidence calibration. \\
\midrule
Internal Field Metrics 
& PathSensLogit / PathSensHidden 
& Differences in output logits or hidden states for the same sample under different reveal paths. \\
\midrule
Internal Field Metrics 
& Trajectory RMSE 
& Root mean squared error between the predicted trajectory and the teacher-provided ground-truth trajectory. \\
\midrule
Internal Field Metrics 
& Derivative RMSE 
& Error between the learned local vector field and the ground-truth vector field. \\
\midrule
Internal Field Metrics 
& Solver Consistency 
& Consistency of terminal states or logits when the same model is evaluated with different integration step sizes or solvers. \\
\midrule
Internal Field Metrics 
& Jacobian Smoothness / JRS 
& Difference between local Jacobians across adjacent layers, or between old and new models in continual learning. \\
\midrule
Internal Field Metrics 
& NormPath / Curvature / VelAlign 
& Measures of hidden trajectory length, bending, and alignment of adjacent velocity directions. \\
\bottomrule
\end{tabularx}
\end{table}

\subsection{PDE Prototype Experiments: From Endpoint Functions to Propagation Generators}
\label{app:pde_prototypes}

The PDE prototype experiments are designed to construct controlled systems with known ground-truth propagation laws, and to compare models that learn only endpoint mappings with models that learn propagation generators. The goal is not to optimize performance on realistic vision tasks, but to turn the question of whether a propagation field is learnable into a controlled numerical experiment. Specifically, we evaluate models in terms of horizon extrapolation, semigroup consistency, step-size reparameterization, and perturbation propagation.

\subsubsection{Data and Dynamical Systems}
\label{app:pde_data}

The input is an initial one-dimensional or low-dimensional field state \(u(x,0)\), generated from combinations of smooth basis functions, localized wave packets, or random initial conditions. A known numerical PDE solver is then used to generate the full trajectory \(u(x,t)\), from which terminal states are sampled at multiple horizons.

We consider three PDE families:

\begin{itemize}[leftmargin=1.5em]
    \item \textbf{PDE-A: Linear advection--diffusion.} 
    This system is used to evaluate smooth propagation, diffusion, and temporal extrapolation.

    \item \textbf{PDE-B: Nonlinear advection--diffusion--reaction.} 
    This system introduces nonlinear reaction terms, source effects, and local perturbation propagation.

    \item \textbf{PDE-C: Non-autonomous variable-coefficient dynamics.} 
    In this system, coefficients vary over time or space, allowing us to test whether models can learn non-stationary propagation laws.
\end{itemize}

\subsubsection{Model Configurations}
\label{app:pde_models}

Table~\ref{tab:pde_models} summarizes the model families used in the PDE prototype experiments.

\begin{table}[t]
\centering
\caption{Model configurations in the PDE prototype experiments.}
\label{tab:pde_models}
\begin{tabularx}{\linewidth}{p{0.25\linewidth} p{0.34\linewidth} X}
\toprule
\textbf{Model} & \textbf{Training Objective} & \textbf{Experimental Role} \\
\midrule
M1: Fixed-Endpoint MLP
& Directly predicts the terminal state at the training horizon from the initial state.
& Tests whether memorizing a fixed endpoint map can extrapolate to other time horizons. \\
\midrule
M2: Time-Conditioned MLP
& Takes the initial state and target time \(T\) as inputs and directly predicts \(u(T)\).
& Tests whether adding time conditioning is sufficient to learn a family of propagation maps. \\
\midrule
M3: Propagation Generator
& Learns a local update rule or vector field and obtains the terminal state through multi-step integration.
& Tests whether the model learns a reparameterizable propagation mechanism rather than a direct endpoint map. \\
\midrule
M4: Structured Propagation Model
& Adds PDE-style inductive structure or stronger propagation constraints on top of the learned generator.
& Tests whether explicit propagation structure improves extrapolation and stability. \\
\bottomrule
\end{tabularx}
\end{table}

\subsubsection{Training and Evaluation Protocol}
\label{app:pde_protocol}

The training and evaluation procedure is as follows:

\begin{enumerate}[leftmargin=1.5em]
    \item Generate training initial conditions, test initial conditions, and ground-truth PDE trajectories across multiple horizons.

    \item Train endpoint models, time-conditioned models, and propagation-generator models on the training horizons.

    \item Evaluate terminal-state prediction at multiple horizons, such as \(T=0.5, 1.0, 1.5,\) and \(2.0\).

    \item For generator-based models, change the numerical integration step size, such as halving \(dt\), and test whether the model preserves the same propagated state.

    \item Add small perturbations to the initial state and compare the model-induced perturbation propagation with the ground-truth PDE perturbation propagation.

    \item Include negative controls, such as random state pairing, shuffled time order, or shuffled trajectories, to test whether the model relies on the true temporal propagation order.
\end{enumerate}

\subsubsection{Metrics and Outputs}
\label{app:pde_metrics}

We use the following metrics to evaluate whether a model learns only endpoint behavior or a stable propagation law:

\begin{itemize}[leftmargin=1.5em]
    \item \textbf{Endpoint MSE:} mean squared error between the predicted terminal state and the ground-truth terminal state.

    \item \textbf{Horizon extrapolation MSE:} terminal-state error evaluated outside the training time horizon.

    \item \textbf{Semigroup consistency:} consistency between directly integrating to \(T\) and composing two shorter integrations, e.g., \(T_1 + T_2 = T\).

    \item \textbf{Regrid \(dt/2\) error:} change in model output when the integration step size is halved.

    \item \textbf{Energy correlation:} correlation between the energy evolution of the model trajectory and that of the ground-truth PDE trajectory.

    \item \textbf{Perturbation MSE:} error between the ground-truth and model-predicted propagation of small initial perturbations.

    \item \textbf{Negative-control gap:} performance gap between the correctly ordered trajectories and randomized or shuffled trajectory controls.
\end{itemize}

\subsection{Propagation-Field Extraction in Ordinary Deep Networks}
\label{app:ordinary_network_fields}

This group of experiments investigates whether the intermediate representations of standard deep architectures, such as ResNets and Transformers, can also be analyzed as trajectories or fields evolving along depth, even in the absence of an explicit PDE teacher. To this end, we treat the layer index as a pseudo-time variable, and record intermediate hidden states, local Jacobian spectra, trajectory curvature, and reparameterization behavior across depth.

\subsubsection{ResNet Continuization and Trajectory Alignment}
\label{app:resnet_continuization}

This experiment examines whether the hidden representations of ResNets exhibit trajectory-like structure when depth is interpreted as pseudo-time.

\begin{itemize}[leftmargin=1.5em]
    \item We train ResNets with different depths, e.g., depth \(=4, 6, 12\).
    \item For the same test samples, we extract the hidden representation after each residual block.
    \item For each sample, the sequence of layerwise hidden states is treated as a trajectory along the depth direction.
    \item We compare the resulting trajectory shapes across networks of different depths using alignment procedures such as PCA-based projection or Procrustes alignment.
    \item We record \textbf{NormPath}, \textbf{Curvature}, \textbf{VelAlign}, and the \textbf{pairwise alignment error across depths}.
\end{itemize}

The purpose of this experiment is to test whether ResNet trajectories admit a continuous-depth interpretation, rather than behaving as unrelated layerwise transformations.

\subsubsection{Comparing Depth Fields in Transformers and ResNets}
\label{app:transformer_resnet_comparison}

This experiment compares the propagation statistics of Transformers and ResNets to determine whether propagation-field phenomena are specific to residual networks or arise more broadly in deep architectures.

\begin{itemize}[leftmargin=1.5em]
    \item We construct Transformers with genuine multi-token inputs rather than degenerate single-token attention.
    \item Intermediate outputs from Transformer blocks and ResNet blocks are mapped into a comparable hidden space.
    \item We compare whether the two architectures exhibit similar depthwise propagation statistics, such as high \textbf{VelAlign} or low \textbf{Curvature}.
    \item The goal is to distinguish whether the propagation-field phenomenon is a special case of residual architecture design or a more general property of deep representation evolution.
\end{itemize}

\subsubsection{Jacobian Spectral Smoothness Analysis}
\label{app:jacobian_smoothness}

This experiment studies whether layerwise transformations evolve smoothly across depth, or instead change abruptly from layer to layer.

\begin{itemize}[leftmargin=1.5em]
    \item For each layer or block, we estimate the local Jacobian spectrum using randomized projections, e.g., Jacobian--vector products (\(Jv\)) or finite-difference approximations.
    \item We compute the Wasserstein distance between the Jacobian spectral distributions of adjacent layers.
    \item We also compute the entropy of the spectral distribution at each layer to examine whether the local propagation operator changes smoothly along depth.
    \item This experiment is designed to test whether inter-layer transformations resemble discrete jumps or the evolution of local operators in a continuous field.
\end{itemize}

\subsubsection{Depth Refinement and Solver Reparameterization}
\label{app:depth_refinement_solver}

This experiment directly evaluates whether layers can be interpreted as numerical slices of a common propagation field.

\begin{itemize}[leftmargin=1.5em]
    \item We fix a learned field in a \textbf{Continuous Flow} or \textbf{Shared-Field} model.
    \item We evaluate multiple discretization depths, e.g., \(8, 16, 32, 64,\) and \(128\) steps.
    \item We also evaluate different numerical solvers, such as Euler, midpoint, and RK4.
    \item Across these discretizations, we compare the terminal hidden-state error, logit error, trajectory error, and prediction agreement.
\end{itemize}

If the model truly represents a coherent propagation field, its predictions should remain stable under refinement of the discretization or changes in the solver.

\subsubsection{Layerwise Comparison of Endpoint Maps, Time-Conditioned Families, and Continuous Flows}
\label{app:layerwise_family_comparison}

This experiment compares three model families under the same task, with the goal of separating \emph{functional equivalence} from \emph{equivalence in internal propagation geometry}. Table~\ref{tab:layerwise_model_families} summarizes the comparison.

\begin{table}[t]
\centering
\caption{Model families compared in the layerwise propagation experiment.}
\label{tab:layerwise_model_families}
\begin{tabularx}{\linewidth}{p{0.28\linewidth} p{0.34\linewidth} X}
\toprule
\textbf{Model Family} & \textbf{Definition} & \textbf{Experimental Role} \\
\midrule
Endpoint Map
& Directly learns a mapping from input to label, without explicitly constraining internal propagation.
& Serves as the endpoint-only baseline. \\
\midrule
Time-Conditioned Family
& Takes time or layer index as an input and learns a family of ordered mappings.
& Tests whether temporal/depth conditioning alone induces structured propagation. \\
\midrule
Continuous Flow
& Explicitly learns a continuous vector field and obtains outputs through numerical integration.
& Represents the strongest propagation-field formulation. \\
\bottomrule
\end{tabularx}
\end{table}

For these three families, we compare:

\begin{itemize}[leftmargin=1.5em]
    \item hidden trajectory length,
    \item curvature,
    \item alignment of velocity directions across depth, and
    \item task accuracy.
\end{itemize}

The purpose of this comparison is to distinguish whether two models that are functionally similar at the endpoint level also exhibit similar internal propagation geometry.

\subsection{Hard-Manifold Classification and True Reparameterization Experiments}
\label{app:hard_manifold_reparameterization}

In the initial experiments, many models achieved high in-distribution accuracy, making it difficult to determine whether propagation-field constraints were genuinely useful. To avoid this saturation effect, we further design harder synthetic manifold classification benchmarks and include OOD evaluation, low-shot regimes, perturbation tests, and true depth-reparameterization experiments.

\subsubsection{Hard-Manifold Data Generation}
\label{app:hard_manifold_data}

We construct multiple low-dimensional curved manifolds embedded in a high-dimensional ambient space. Each class corresponds to a manifold segment characterized by a class center, subspace directions, and nonlinear deformation terms. Task difficulty is controlled by the distance between class centers, noise magnitude, subspace overlap, and manifold curvature.

We consider three difficulty levels: \textbf{easy}, \textbf{medium}, and \textbf{hard}. These settings are designed to prevent all models from reaching saturated accuracy and to make the task sensitive to both functional expressivity and internal propagation structure.

OOD evaluation is constructed through several controlled distribution shifts:

\begin{itemize}[leftmargin=1.5em]
    \item stronger input noise,
    \item rotations of the underlying subspaces,
    \item shifts of class centers, and
    \item changes in the nonlinear curvature terms.
\end{itemize}

We also evaluate low-shot learning regimes with \(20\), \(50\), and \(100\) samples per class.

\subsubsection{Models and Fairness Controls}
\label{app:hard_manifold_models}

We compare several model families, including \textbf{Endpoint}, \textbf{TimeCond}, \textbf{SharedField}, and \textbf{Continuous} models. Whenever possible, hidden dimensions and parameter counts are controlled to make the comparison fair across architectures.

All models are trained using the same optimizer, number of training epochs, batch size, and data splits. For each difficulty level and random seed, we record both functional metrics and geometric metrics. The purpose of this experiment is not to rank models solely by accuracy, but to examine whether geometric smoothness has a conditional relationship with OOD robustness and low-shot generalization.

\subsubsection{Reparameterization Experiments}
\label{app:true_reparameterization}

To test whether a learned model can be interpreted as a discretization of a common propagation field, we first train a single learned field and then vary the number of integration steps and the numerical solver only at test time.

Importantly, we do not retrain separate networks at different depths. This avoids conflating ``different models'' with ``different numerical slices of the same field.'' Instead, the same learned field is evaluated under different discretization schemes.

We record the following metrics:

\begin{itemize}[leftmargin=1.5em]
    \item \textbf{PredAgreeRef:} prediction agreement with a reference discretization;
    \item \textbf{FinalHiddenErr:} error in the terminal hidden state relative to the reference solution;
    \item \textbf{TrajectoryErr:} discrepancy between trajectories under different discretizations;
    \item \textbf{LogitErr:} discrepancy between output logits under different solvers or step sizes.
\end{itemize}

If the errors decrease as the number of integration steps increases or as the solver order improves, the result supports the interpretation that the learned representation behaves like a numerically discretized propagation field.

\subsection{Controlled Teacher-Flow Experiments: Endpoint Function or Propagation Field?}
\label{app:teacher_flow}

The teacher-flow experiments construct a ground-truth latent ODE in which both the true trajectory and the true derivative field are accessible. Their central purpose is to remove the ambiguity present in real tasks, where the true internal field is unobserved, and to directly test whether endpoint supervision is sufficient to recover the underlying propagation field.

\subsubsection{Teacher Dynamics}
\label{app:teacher_dynamics}

We define a latent state \(z(t)\) in a two-dimensional or low-dimensional space. The ground-truth teacher vector field is denoted by
\begin{equation}
\frac{d z(t)}{dt} = f^\star(z(t), t),
\end{equation}
where \(f^\star\) includes rotational, contractive, and nonlinear components. Starting from randomly sampled initial states \(z(0)\), we numerically integrate the teacher dynamics to obtain the full latent trajectory
\begin{equation}
\{z(t_0), z(t_1), \ldots, z(t_K)\}.
\end{equation}
The observed input \(x\) is then generated from the latent initial state through a fixed observation map,
\begin{equation}
x = \psi(z(0)).
\end{equation}
This construction makes the latent trajectory and derivative field available for evaluation, while the model receives only the observed input and task labels during standard endpoint-supervised training.

\subsubsection{Task Definitions}
\label{app:teacher_tasks}

We consider two task types, summarized in Table~\ref{tab:teacher_tasks}.

\begin{table}[t]
\centering
\caption{Task definitions in the teacher-flow experiments.}
\label{tab:teacher_tasks}
\begin{tabularx}{\linewidth}{p{0.28\linewidth} p{0.34\linewidth} X}
\toprule
\textbf{Task} & \textbf{Label Definition} & \textbf{Experimental Role} \\
\midrule
Task A: Endpoint Label
& The label is determined only by the region, angle, or a linear classifier applied to the terminal state \(z(T)\).
& Tests whether supervision on endpoint labels is sufficient to recover the true propagation field. \\
\midrule
Task B: Path Label
& The label is determined by a path-dependent quantity, such as a trajectory integral, winding direction, or path event.
& Tests whether path-dependent labels provide stronger identifiability of the propagation field. \\
\bottomrule
\end{tabularx}
\end{table}

\subsubsection{Models and Losses}
\label{app:teacher_models_losses}

We compare three model classes:

\begin{itemize}[leftmargin=1.5em]
    \item \textbf{M1: EndpointMLP.} 
    This model directly predicts the task label from the observed input \(x\), without explicitly modeling latent trajectories.

    \item \textbf{M2: FlowClassifier with endpoint-only loss.} 
    This model uses a flow-based architecture, but is trained only with endpoint label supervision.

    \item \textbf{M3: FlowClassifier with field losses.} 
    This model uses the same flow-based architecture as M2, but augments the endpoint loss with trajectory loss, derivative loss, and solver-consistency loss.
\end{itemize}

The key controlled comparison is between M2 and M3: they use the same model class, but differ only in the training objective. This allows us to isolate whether endpoint supervision alone is sufficient for field recovery.

The endpoint-only objective is
\begin{equation}
\mathcal{L}_{\mathrm{endpoint}}
=
\mathcal{L}_{\mathrm{task}}.
\end{equation}

The field-aware objective is
\begin{equation}
\mathcal{L}_{\mathrm{field}}
=
\mathcal{L}_{\mathrm{task}}
+
\alpha \mathcal{L}_{\mathrm{traj}}
+
\beta \mathcal{L}_{\mathrm{deriv}}
+
\gamma \mathcal{L}_{\mathrm{solver}},
\end{equation}
where
\begin{equation}
\mathcal{L}_{\mathrm{traj}}
=
\sum_{k=1}^{K}
\|\hat z(t_k) - z(t_k)\|_2^2,
\end{equation}
and
\begin{equation}
\mathcal{L}_{\mathrm{deriv}}
=
\sum_{k=1}^{K}
\|
f_\theta(\hat z(t_k),t_k) - f^\star(z(t_k),t_k)
\|_2^2.
\end{equation}
The solver-consistency term \(\mathcal{L}_{\mathrm{solver}}\) measures whether the same learned field produces consistent predictions under different numerical solvers or integration step sizes.

\subsubsection{Evaluation Metrics}
\label{app:teacher_metrics}

We evaluate both endpoint task performance and internal field recovery:

\begin{itemize}[leftmargin=1.5em]
    \item \textbf{Task Accuracy:} whether the model predicts the correct class label.

    \item \textbf{Trajectory RMSE:} root mean squared error between the predicted trajectory and the teacher trajectory.

    \item \textbf{Derivative RMSE:} error between the learned vector field and the teacher vector field.

    \item \textbf{Reparameterization Consistency:} output consistency of the same learned field under different solvers or integration step counts.

    \item \textbf{NormPath / Curvature / VelAlign / JacWDist:} geometric measures of internal field quality, including trajectory length, bending, velocity alignment, and Wasserstein distance between local Jacobian spectra.
\end{itemize}

\subsection{Reveal-Path Tasks and Real/Semi-Real Task Transfer Experiments}
\label{app:reveal_path_transfer}

This group of experiments transfers the propagation-field idea from controlled teacher-flow settings to scenarios closer to real machine learning tasks. The core idea is that the same input object can be gradually revealed to a model through different observation paths. If the label is invariant to the reveal path, an ideal model should form path-consistent terminal representations and predictions.

\subsubsection{Reveal-Path Construction}
\label{app:reveal_path_construction}

We construct reveal paths for image, text, and speech tasks. Each reveal path defines an ordered sequence of partial observations of the same underlying input. Table~\ref{tab:reveal_path_construction} summarizes the path types used across modalities.

\begin{table}[t]
\centering
\caption{Reveal-path construction across modalities.}
\label{tab:reveal_path_construction}
\begin{tabularx}{\linewidth}{p{0.22\linewidth} p{0.34\linewidth} X}
\toprule
\textbf{Task Type} & \textbf{Reveal-Path Examples} & \textbf{Description} \\
\midrule
Image Tasks
& Center, rows, spiral, frequency.
& The same image is gradually presented through different spatial or frequency-domain paths. \\
\midrule
Text Tasks
& Prefix reveal, sentence-level reveal, evidence-fragment reveal.
& The same text or semantic evidence is exposed in different orders. \\
\midrule
Speech Tasks
& Temporal windows, masked chunks, progressive frequency-band reveal.
& The same speech command is gradually provided through different temporal or time-frequency windows. \\
\bottomrule
\end{tabularx}
\end{table}

\subsubsection{Objective Ablations}
\label{app:reveal_objective_ablations}

We compare several training objectives to isolate which component of the field-aware objective contributes to path consistency, numerical stability, or generalization. Table~\ref{tab:objective_ablations} summarizes the ablation design.

\begin{table}[t]
\centering
\caption{Objective ablations for reveal-path and field-aware training.}
\label{tab:objective_ablations}
\begin{tabularx}{\linewidth}{p{0.18\linewidth} p{0.38\linewidth} X}
\toprule
\textbf{Objective} & \textbf{Loss Components} & \textbf{Experimental Purpose} \\
\midrule
\texttt{task}
& Task loss only, such as classification loss.
& Serves as the standard endpoint-supervision baseline. \\
\midrule
\texttt{reveal}
& Task loss plus hidden-state or logit consistency across reveal paths.
& Tests whether path consistency improves unseen-path generalization or OOD robustness. \\
\midrule
\texttt{solver}
& Task loss plus terminal-state consistency across different step sizes or numerical solvers.
& Tests whether numerical reparameterization stability is beneficial. \\
\midrule
\texttt{jac}
& Task loss plus local Jacobian smoothness or spectral smoothness regularization.
& Tests whether smoothing local propagation operators improves internal field quality or generalization. \\
\midrule
\texttt{full}
& Task loss plus reveal consistency, solver consistency, and Jacobian regularization.
& Tests whether the complete field-aware objective improves both internal propagation quality and downstream generalization. \\
\bottomrule
\end{tabularx}
\end{table}

\subsubsection{Real and Semi-Real Task Transfer}
\label{app:real_task_transfer}

We evaluate reveal-path and field-aware objectives on both semi-real and real tasks:

\begin{itemize}[leftmargin=1.5em]
    \item \textbf{Semi-real task:} reveal-path experiments on \texttt{sklearn} Digits.
    \item \textbf{Real vision tasks:} CIFAR-100 with patch reveal, and Tiny-ImageNet with patch or frequency reveal.
    \item \textbf{Real text tasks:} AG News and SST-2 with prefix reveal.
    \item \textbf{Real speech task:} Speech Commands with temporal reveal.
\end{itemize}

Across all tasks, we record a unified set of metrics: \textbf{AccAvg}, \textbf{UnseenAcc}, \textbf{OODAcc}, \textbf{PathSensLogit}, \textbf{PathSensHidden}, \textbf{cross-path agreement}, and \textbf{ECE}. These metrics jointly evaluate endpoint accuracy, unseen-path generalization, robustness to distributional shifts, internal path sensitivity, prediction consistency, and calibration.

\subsubsection{Collapse Prevention and Diagnostic Metrics}
\label{app:collapse_prevention}

Path consistency alone does not imply a good model. A degenerate classifier that always predicts a constant class can also achieve low path sensitivity. Therefore, when using field-aware objectives, we jointly monitor accuracy, prediction entropy, class balance, calibration, and cross-path agreement.

This distinction is important because a reduction in path sensitivity can arise from two qualitatively different mechanisms. In the desirable case, the model learns informative path-invariant representations. In the degenerate case, the model collapses to uninformative representations or constant predictions. We therefore interpret path-consistency improvements only together with endpoint performance and anti-collapse diagnostics.

\subsection{Training-Algorithm Pareto Analysis and ``Anti-Field'' Dynamics}
\label{app:training_pareto_antifield}

Once propagation-field quality is included in the training objective, optimization is no longer governed by a single task loss. Instead, it becomes a multi-objective problem involving task performance, field quality, and computational cost. This group of experiments compares the Pareto behavior of different training algorithms under the same field model class and the same teacher-flow environment.

\subsubsection{Compared Algorithms}
\label{app:compared_training_algorithms}

Table~\ref{tab:training_algorithms} summarizes the training algorithms compared in this experiment.

\begin{table}[t]
\centering
\caption{Training algorithms for task--field--compute Pareto analysis.}
\label{tab:training_algorithms}
\begin{tabularx}{\linewidth}{p{0.22\linewidth} p{0.38\linewidth} X}
\toprule
\textbf{Algorithm} & \textbf{Training Strategy} & \textbf{Question Tested} \\
\midrule
FullBPTT
& End-to-end backpropagation through the full solver.
& Tests whether the standard training method still lies on the Pareto frontier. \\
\midrule
SegmentShooting
& Segment-wise shooting or collocation with consistency constraints between local segments.
& Tests whether local dynamical training improves field recovery. \\
\midrule
LocalFieldMatch
& Directly matches the local vector field or derivatives.
& Tests whether local field supervision is more effective than global endpoint supervision. \\
\midrule
PCGrad / MGDA
& Applies multi-objective gradient-conflict mitigation.
& Tests whether conflicts between task and field gradients require specialized optimization methods. \\
\midrule
Curriculum
& Trains the field or local dynamics first, then trains the task objective.
& Tests whether training order changes the resulting Pareto point. \\
\midrule
Alternating
& Alternates between task-optimization steps and field-optimization steps.
& Tests whether alternating optimization improves the trade-off among objectives. \\
\midrule
ProjectedTask
& Projects the task gradient onto directions that do not harm the field objective.
& Tests whether ``anti-field'' conflicts can be mitigated through gradient intervention. \\
\bottomrule
\end{tabularx}
\end{table}

\subsubsection{Training-Dynamics Logging}
\label{app:training_dynamics_logging}

For each training epoch, we record task accuracy, trajectory RMSE, derivative RMSE, and reparameterization error. We also compute the cosine similarity between the task gradient and the field gradient:
\begin{equation}
\cos(g_{\mathrm{task}}, g_{\mathrm{field}})
=
\frac{
\langle g_{\mathrm{task}}, g_{\mathrm{field}} \rangle
}{
\|g_{\mathrm{task}}\|_2 \|g_{\mathrm{field}}\|_2
}.
\end{equation}

We report the mean cosine similarity, the minimum cosine similarity, and the fraction of epochs or optimization steps with negative cosine values. These statistics are used to determine whether task--field conflicts emerge in later stages of training. Finally, we compare the algorithms along three axes: task performance, field recovery, and computational cost, thereby characterizing their task--field--compute Pareto trade-offs.
\subsection{Field Preservation Regularization (FPR) in Continual Learning}
\label{app:fpr_continual_learning}

The FPR experiments apply the propagation-field perspective to continual learning. The central question is whether catastrophic forgetting can be interpreted not only as degradation of endpoint performance, but also as the rewriting of internal propagation fields associated with previous tasks. We therefore ask whether preserving the hidden trajectories and local Jacobian fields of old tasks can reduce forgetting or provide an additional dimension of retention beyond accuracy.

\subsubsection{Dataset and Task Split}
\label{app:fpr_dataset_split}

We use Split CIFAR-100 as the continual learning benchmark. The dataset is divided into \(20\) sequential tasks, with each task containing \(5\) classes. Models are trained task by task. After finishing task \(t\), we evaluate the model on all tasks observed so far.

The baseline diagnostic setting is pure sequential fine-tuning, without replay and without regularization-based protection. The final outputs of the experimental pipeline include the accuracy matrix, drift records, correlation reports, and phase-wise comparison tables.

\subsubsection{Model: FieldResNet-18}
\label{app:fieldresnet18}

The backbone is a ResNet-18-style architecture, modified to explicitly expose hidden states from multiple intermediate layers. In our experiments, we record \(6\) intermediate layers covering early, middle, and late depths of the network.

For each anchor sample, we record the hidden trajectory and a local Jacobian proxy under both the old model and the new model. This allows us to compare how the same input is propagated internally before and after subsequent task learning.

\subsubsection{Compared Methods}
\label{app:fpr_compared_methods}

Table~\ref{tab:fpr_methods} summarizes the compared continual learning methods.

\begin{table}[t]
\centering
\caption{Compared methods in the FPR continual learning experiments.}
\label{tab:fpr_methods}
\begin{tabularx}{\linewidth}{p{0.20\linewidth} p{0.42\linewidth} X}
\toprule
\textbf{Method} & \textbf{Core Mechanism} & \textbf{Notes} \\
\midrule
Finetune
& Sequential training without protection.
& Forgetting baseline. \\
\midrule
EWC
& Fisher-weighted parameter-space \(L_2\) regularization.
& Protects parameters. \\
\midrule
LwF
& Distills logits from the old model.
& Protects functional outputs, with no or weak sample storage. \\
\midrule
ER
& Stores a sample buffer and trains with cross-entropy on mixed old and new samples.
& Replay baseline. \\
\midrule
DER++
& Stores samples, old logits, and labels; aligns old logits using MSE together with cross-entropy.
& Strong replay/distillation baseline. \\
\midrule
FPR-Traj
& Preserves hidden trajectories of the old and new models on anchor samples.
& Constrains state propagation only. \\
\midrule
FPR-Jac
& Preserves local Jacobian fields of the old and new models.
& Constrains local propagation operators only. \\
\midrule
FPR-Full
& Preserves both hidden trajectories and local Jacobian fields.
& Full field-preservation variant. \\
\bottomrule
\end{tabularx}
\end{table}

\subsubsection{Key Evaluation Metrics}
\label{app:fpr_metrics}

Table~\ref{tab:fpr_metrics} summarizes the key evaluation metrics.

\begin{table}[t]
\centering
\caption{Evaluation metrics for FPR continual learning experiments.}
\label{tab:fpr_metrics}
\begin{tabularx}{\linewidth}{p{0.22\linewidth} p{0.38\linewidth} X}
\toprule
\textbf{Metric} & \textbf{Definition} & \textbf{Interpretation} \\
\midrule
AA
& Average accuracy of the final model over all tasks.
& Overall continual learning performance. \\
\midrule
BWT
& Change in old-task accuracy from immediately after learning the task to the final model.
& Negative values indicate forgetting; values closer to \(0\) are better. \\
\midrule
FWT
& Initial performance on a task before learning it, relative to a random baseline.
& Measures forward transfer. \\
\midrule
FRS
& Difference between hidden trajectories of the old and new models on anchor samples.
& Lower values indicate better representation-trajectory retention. \\
\midrule
JRS
& Difference between local Jacobian proxies of the old and new models on anchor samples.
& Lower values indicate better retention of local propagation operators. \\
\midrule
FRS\_final / JRS\_final
& Field-retention metrics comparing the final model with previous task-specific models.
& Used to analyze long-term field retention. \\
\bottomrule
\end{tabularx}
\end{table}

\subsubsection{Phase 0: Pure Forgetting Diagnostics}
\label{app:fpr_phase0}

In Phase 0, we train the model sequentially on all \(20\) tasks using the Finetune baseline. After each task is completed, we evaluate the model on all previously observed tasks and generate an accuracy matrix.

For each old-task anchor sample, we record parameter drift, trajectory drift, Jacobian drift, and accuracy drop. The phase outputs include \texttt{accuracy\_matrix.json}, \texttt{drift\_records.csv}, \texttt{correlation\_report.json}, \texttt{p0\_drift\_heatmap.png}, and \texttt{p0\_drift\_scatter.png}. The goal of this phase is diagnostic: to determine which types of drift are more closely associated with forgetting, rather than to propose a new method.

\subsubsection{Phase 1: Full Comparison of Eight Methods}
\label{app:fpr_phase1}

In Phase 1, we run Finetune, EWC, LwF, ER, DER++, FPR-Traj, FPR-Jac, and FPR-Full under the same task order, model architecture, and evaluation protocol. We record the complete accuracy matrix and report AA, BWT, FWT, FRS, and JRS.

The phase outputs include \texttt{results.json} and \texttt{p1\_comparison.png}. This phase compares traditional continual learning methods and FPR variants in terms of both function retention and field retention.

\subsubsection{Phase 2: FPR Component and Layer Ablations}
\label{app:fpr_phase2}

Phase 2 performs component and layer-level ablations. The component ablation compares TrajOnly, JacOnly, and Full variants. The layer ablation compares field preservation applied to early layers, middle layers, late layers, and all selected layers.

We compare the effects of different FPR components and protected layer groups on AA, BWT, FRS, and JRS. The phase outputs include \texttt{p2\_ablation.png} and \texttt{results.json}. This phase is used to determine which objects and which layers should be protected for effective propagation-field retention.

\subsubsection{Phase 3: Budget-Efficiency Pareto Experiments}
\label{app:fpr_phase3}

Phase 3 studies the budget-efficiency trade-off. We vary the anchor or memory budget, for example \(50\), \(100\), \(200\), and \(500\) samples per task. Under each budget, we compare Finetune, ER, DER++, and FPR-Full using AA, BWT, FRS, and JRS.

The phase outputs include \texttt{budget\_results.json} and \texttt{p3\_budget.png}. This phase examines whether FPR can provide advantages under low-budget settings, especially when label storage or replay memory is limited, and whether it is complementary to replay-based methods.

\subsubsection{Phase 4: Hybrid Enhancement Experiments}
\label{app:fpr_phase4}

In Phase 4, FPR is no longer treated only as a standalone continual learning method. Instead, it is used as a field-retention regularizer that can be added to existing continual learning methods.

We compare ER with ER+FPR-Late, and DER++ with DER+++FPR-Full. Finetune and FPR-Full are retained as the forgetting lower bound and the standalone FPR baseline, respectively. All methods are run under the same memory or anchor budget, with particular focus on small to medium budgets such as \(50\), \(100\), and \(200\) samples per task.

We record AA, BWT, FWT, FRS, JRS, FRS\_final, and JRS\_final. For each hybrid method, we also compute its improvement relative to the corresponding base method:
\begin{equation}
\Delta \mathrm{AA}
=
\mathrm{AA}_{\mathrm{hybrid}}
-
\mathrm{AA}_{\mathrm{base}},
\end{equation}
\begin{equation}
\Delta \mathrm{BWT}
=
\mathrm{BWT}_{\mathrm{hybrid}}
-
\mathrm{BWT}_{\mathrm{base}},
\end{equation}
\begin{equation}
\Delta \mathrm{FRS}
=
\mathrm{FRS}_{\mathrm{hybrid}}
-
\mathrm{FRS}_{\mathrm{base}},
\end{equation}
and
\begin{equation}
\Delta \mathrm{JRS}
=
\mathrm{JRS}_{\mathrm{hybrid}}
-
\mathrm{JRS}_{\mathrm{base}}.
\end{equation}
This phase evaluates whether field preservation can improve existing replay or distillation methods by adding an internal-propagation retention objective.

\subsection{Additional Experiment Result}

\begin{table*}[t]
\centering
\caption{
\textbf{PDE prototype experiments.}
Controlled PDE systems with known propagation laws are used to compare endpoint maps, time-conditioned maps, propagation generators, and structured propagation models. Metrics test horizon extrapolation, step-size reparameterization, energy evolution, and perturbation propagation. Lower MSE/regrid/perturbation error is better; higher energy correlation is better.
}
\label{tab:app_pde_summary}
\scriptsize
\setlength{\tabcolsep}{4pt}
\renewcommand{\arraystretch}{1.05}
\begin{adjustbox}{width=\textwidth}
\begin{tabular}{llccccc}
\toprule
\textbf{System} & \textbf{Model} & \textbf{Endpoint MSE} & \textbf{\(T=2.0\) MSE} & \textbf{Regrid \(dt/2\)} & \textbf{Energy \(r\)} & \textbf{Perturb. MSE} \\
\midrule
\multirow{4}{*}{PDE-A: Linear Adv-Diff}
& M1 Endpoint & 2.529e-03 & 1.217e+00 & \na & \na & \na \\
& M2 TimeCond & 2.506e-01 & 1.394e+00 & \na & \na & \na \\
& M3 Generator & 8.220e-01 & 1.933e+00 & \best{1.772e-03} & 0.986 & \best{7.287e-03} \\
& M4 Structured & 1.748e-01 & \best{1.948e-01} & 2.170e-02 & \best{0.995} & 7.628e-03 \\
\midrule
\multirow{4}{*}{PDE-B: Nonlinear ADR}
& M1 Endpoint & 4.638e-03 & 4.056e-01 & \na & \na & \na \\
& M2 TimeCond & 8.382e-02 & 7.337e-01 & \na & \na & \na \\
& M3 Generator & 3.214e-01 & 7.846e-01 & \best{6.558e-03} & \best{0.998} & \best{5.547e-03} \\
& M4 Structured & 1.174e-01 & \best{9.944e-02} & 2.980e-02 & 0.976 & 7.500e-03 \\
\midrule
\multirow{4}{*}{PDE-C: Non-autonomous}
& M1 Endpoint & 3.832e-03 & 5.024e-01 & \na & \na & \na \\
& M2 TimeCond & 1.113e-01 & 7.990e-01 & \na & \na & \na \\
& M3 Generator & 5.598e-01 & 9.267e-01 & \best{8.443e-03} & 0.977 & \best{8.817e-03} \\
& M4 Structured & 1.260e-01 & \best{9.195e-02} & 2.494e-02 & \best{0.987} & 9.372e-03 \\
\midrule
\multirow{3}{*}{PDE-B controls}
& M3 NC1 rand-pair & 1.712e-01 & \na & 1.703e-01 & \na & \na \\
& M3 NC2 shuffled & 1.376e+00 & \na & 1.391e+00 & \na & \na \\
& M3 normal & 3.214e-01 & \na & 3.207e-01 & \na & \na \\
\bottomrule
\end{tabular}
\end{adjustbox}
\vspace{0.25em}
\footnotesize
\emph{Message:} endpoint fitting can match a training horizon, but stable propagation recovery requires generator-style or structured propagation models.
\end{table*}

\begin{table*}[t]
\centering
\caption{
\textbf{Propagation-field extraction and controlled teacher-flow experiments.}
Ordinary networks are analyzed by treating depth as pseudo-time. Teacher-flow experiments use a latent ODE with accessible ground-truth trajectories and derivatives, allowing direct comparison between endpoint accuracy and field recovery.
}
\label{tab:app_field_extraction_teacher}
\scriptsize
\setlength{\tabcolsep}{4pt}
\renewcommand{\arraystretch}{1.05}
\begin{adjustbox}{width=\textwidth}
\begin{tabular}{llcccccc}
\toprule
\textbf{Block} & \textbf{Setting / Model} & \textbf{Acc.} & \textbf{NormPath} & \textbf{Curv.} & \textbf{VelAlign} & \textbf{Traj./Spectral} & \textbf{Deriv./Jac.} \\
\midrule
\multirow{3}{*}{ResNet depth}
& Depth 4 & 1.0000 & 1.1936 & 0.0480 & 0.9873 & \na & \na \\
& Depth 6 & 1.0000 & 1.2982 & 0.0822 & 0.9705 & \na & \na \\
& Depth 12 & 1.0000 & 1.4656 & 0.2177 & 0.9173 & \na & \na \\
\midrule
\multirow{2}{*}{Architecture}
& ResNet & 1.0000 & 1.2803 & 0.0885 & 0.9702 & \na & \na \\
& Transformer & 1.0000 & 1.3418 & 0.0816 & 0.9709 & Procrustes 1.3968 & \na \\
\midrule
\multirow{2}{*}{Jacobian spectra}
& Adjacent-layer WDist & \na & \na & \na & \na & mean 0.00833 & min 0.0029 / max 0.0168 \\
& Depth refinement & 1.0000 & \na & \na & \na & hidden/logit err. \(10^{-6}\)--\(10^{-5}\) & \na \\
\midrule
\multirow{3}{*}{Teacher A}
& M1 EndpointMLP & 0.9975 & \na & \na & \na & \na & \na \\
& M2 FlowEndOnly & 0.9975 & 6.5657 & \na & \na & Traj. 3.5229 & Deriv. 2.7627 / JacW 0.2505 \\
& M3 FlowFieldLoss & 0.9975 & \best{1.6595} & \na & \na & \best{Traj. 0.0590} & \best{Deriv. 0.1495 / JacW 0.0409} \\
\midrule
\multirow{3}{*}{Teacher B}
& M1 EndpointMLP & 0.9937 & \na & \na & \na & \na & \na \\
& M2 FlowEndOnly & 0.9900 & 4.0904 & \na & \na & Traj. 3.3263 & Deriv. 1.8882 / JacW 0.2170 \\
& M3 FlowFieldLoss & 0.9900 & \best{1.6551} & \na & \na & \best{Traj. 0.0694} & \best{Deriv. 0.1632 / JacW 0.0333} \\
\bottomrule
\end{tabular}
\end{adjustbox}
\vspace{0.25em}
\footnotesize
\emph{Message:} ordinary deep networks expose measurable depth-wise geometry; in teacher-flow systems, identical endpoint accuracy can hide large differences in true field recovery.
\end{table*}

\begin{table*}[t]
\centering
\caption{
\textbf{Real and semi-real reveal-path transfer.}
The same object is revealed through multiple paths: image patch/frequency reveal, text prefix/evidence reveal, and speech temporal-window reveal. Metrics jointly test endpoint accuracy, unseen-path generalization, OOD robustness, path sensitivity, and calibration.
}
\label{tab:app_real_transfer}
\scriptsize
\setlength{\tabcolsep}{4pt}
\renewcommand{\arraystretch}{1.05}
\begin{adjustbox}{width=\textwidth}
\begin{tabular}{llcccccc}
\toprule
\textbf{Dataset} & \textbf{Obj.} & \textbf{AccAvg} & \textbf{Unseen} & \textbf{OOD} & \textbf{PathSensLogit} & \textbf{PathSensHidden} & \textbf{ECE} \\
\midrule
\multirow{3}{*}{digits}
& task & \best{0.7528} & \best{0.6633} & \best{0.8089} & 8.4402 & 6.2641 & \best{0.0325} \\
& reveal & 0.5261 & 0.4222 & 0.5978 & \best{3.5193} & 2.1058 & 0.1746 \\
& full & 0.1894 & 0.1833 & 0.1844 & 10.4550 & \best{2.0529} & 0.5876 \\
\midrule
\multirow{3}{*}{CIFAR-100}
& task & 0.0405 & 0.0295 & 0.0570 & 15.1867 & 7.2761 & 0.0473 \\
& reveal & \best{0.0558} & \best{0.0490} & \best{0.0600} & 7.2521 & 3.3842 & 0.0239 \\
& full & 0.0093 & 0.0090 & 0.0100 & \best{3.0355} & \best{2.1391} & \best{0.0229} \\
\midrule
\multirow{3}{*}{Tiny-ImageNet}
& task & 0.0970 & 0.0853 & 0.1000 & 4.4924 & 4.4374 & 0.1018 \\
& reveal & 0.0943 & 0.0853 & 0.1053 & 1.7353 & 2.3665 & 0.0539 \\
& full & \best{0.1180} & \best{0.1087} & \best{0.1280} & \best{1.2742} & \best{2.0651} & \best{0.0182} \\
\midrule
\multirow{3}{*}{AG News}
& task & \best{0.3642} & \best{0.3639} & \best{0.2753} & \best{0.0470} & \best{0.1387} & \best{0.0073} \\
& reveal & 0.2804 & 0.2802 & 0.2603 & 12.0342 & 4.0788 & 0.0337 \\
& full & 0.2500 & 0.2500 & 0.2505 & 0.1302 & 0.1644 & 0.2182 \\
\midrule
\multirow{3}{*}{SST-2}
& task & \best{0.5789} & \best{0.5792} & 0.5261 & \best{0.0078} & 0.1158 & \best{0.0045} \\
& reveal & 0.5379 & 0.5380 & 0.5409 & 4.3906 & 0.4716 & 0.0250 \\
& full & 0.5572 & 0.5572 & \best{0.5572} & 0.0592 & \best{0.1010} & 0.0992 \\
\midrule
\multirow{3}{*}{Speech}
& task & \best{0.1925} & \best{0.1917} & 0.1298 & 0.2381 & 0.3804 & 0.0465 \\
& reveal & 0.1866 & \best{0.1917} & \best{0.1386} & 0.1071 & 0.1591 & 0.0547 \\
& full & 0.1121 & 0.1121 & 0.1091 & \best{0.0682} & \best{0.1331} & \best{0.0356} \\
\bottomrule
\end{tabular}
\end{adjustbox}
\vspace{0.25em}
\footnotesize
\emph{Message:} field-aware objectives help when aligned with task structure, but over-regularization can reduce PathSens while collapsing accuracy.
\end{table*}

\begin{table*}[t]
\centering
\caption{
\textbf{Current strict-suite summary: task--field--compute tradeoffs and regime dependence.}
For task-structure and phase-diagram rows, \(\Delta\) denotes the field-aware objective minus the task-only baseline; negative \(\Delta\)Field means lower field error. For Pareto rows, Task is TestAcc and Field reports Traj./Deriv. RMSE. The results show that field-aware learning is useful only in specific path-structured regimes and that optimization methods occupy different task--field--compute tradeoff points.
}
\label{tab:app_strict_suite_summary}
\scriptsize
\setlength{\tabcolsep}{3.5pt}
\renewcommand{\arraystretch}{1.08}
\begin{adjustbox}{width=\textwidth}
\begin{tabular}{llccccc}
\toprule
\textbf{Block} & \textbf{Setting / Method} & \textbf{Task Axis} & \textbf{OOD Axis} & \textbf{Field Axis} & \textbf{Conflict / Selector} & \textbf{Takeaway} \\
\midrule
\multirow{4}{*}{Task structure}
& \(\lambda=0.0\) endpoint
& \(\Delta\)ID \(-0.530\)
& \(\Delta\)OOD \(-0.115\)
& \(\Delta\)Traj \(-1.925\) / \(\Delta\)Deriv \(-1.830\)
& --
& field improves, task collapses \\

& \(\lambda=0.5\) medium path
& \(\Delta\)ID \(-0.070\)
& \(\Delta\)OOD \(+0.210\)
& \(\Delta\)Traj \(-0.811\) / \(\Delta\)Deriv \(-0.031\)
& --
& OOD gain despite ID drop \\

& \(\lambda=0.75\) strong path
& \(\Delta\)ID \(+0.015\)
& \(\Delta\)OOD \(+0.075\)
& \(\Delta\)Traj \(-0.260\) / \(\Delta\)Deriv \(-0.722\)
& --
& best aligned regime \\

& \(\lambda=1.0\) long path
& \(\Delta\)ID \(-0.185\)
& \(\Delta\)OOD \(-0.030\)
& \(\Delta\)Traj \(-0.840\) / \(\Delta\)Deriv \(-0.643\)
& --
& over-constrained \\
\midrule
\multirow{4}{*}{Pareto training}
& FullBPTT
& Acc. \(0.590\)
& --
& Traj \(1.053\) / Deriv \(2.674\)
& neg-grad \(0.18\)
& baseline point \\

& LocalFieldMatch
& Acc. \(0.615\)
& --
& Traj \(1.034\) / Deriv \(2.655\)
& neg-grad \(0.36\)
& better field and accuracy \\

& MGDA
& Acc. \(0.665\)
& --
& Traj \(1.297\) / Deriv \(2.830\)
& neg-grad \(0.55\)
& best task accuracy \\

& Curriculum
& Acc. \(0.660\)
& --
& Traj \(1.115\) / Deriv \(2.735\)
& neg-grad \(0.00\)
& stable compromise \\
\midrule
\multirow{2}{*}{Anti-field dynamics}
& FullBPTT
& Acc. \(0.590\)
& --
& Traj \(1.053\) / Deriv \(2.674\)
& cos \(0.405\), neg \(0.18\)
& task--field conflict emerges \\

& ProjectedTask
& Acc. \(0.670\)
& --
& Traj \(1.305\) / Deriv \(2.831\)
& cos \(-0.025\), neg \(0.55\)
& task improves, field worsens \\
\midrule
\multirow{3}{*}{Phase diagram}
& \(\lambda=0.0\), avg.
& \(\Delta\)ID \(-0.329\)
& \(\Delta\)OOD \(-0.323\)
& \(\Delta\)Traj \(-0.660\) / \(\Delta\)Deriv \(-0.219\)
& joint \(0/16\)
& endpoint-like over-regularization \\

& \(\lambda=0.5\), avg.
& \(\Delta\)ID \(+0.137\)
& \(\Delta\)OOD \(+0.064\)
& \(\Delta\)Traj \(-0.567\) / \(\Delta\)Deriv \(-0.088\)
& joint \(8/16\)
& main positive window \\

& \(\lambda=1.0\), avg.
& \(\Delta\)ID \(-0.031\)
& \(\Delta\)OOD \(-0.087\)
& \(\Delta\)Traj \(-0.190\) / \(\Delta\)Deriv \(-0.034\)
& joint \(0/16\)
& strong path alone insufficient \\
\midrule
\multirow{2}{*}{Multivariate}
& Global predictors of \(\Delta\)OOD
& corr\((\lambda,\Delta\mathrm{OOD})=0.461\)
& --
& corr\((\Delta\mathrm{Deriv},\Delta\mathrm{OOD})=0.556\)
& RF: Deriv \(0.363\), \(\lambda\) \(0.347\), Jac \(0.210\)
& path structure and local field matter \\

& Stratified field proxy
& --
& low-\(\lambda\): \(-0.33/-0.34\)
& high-\(\lambda\): \(+0.32/+0.29\)
& --
& field quality is conditionally useful \\
\bottomrule
\end{tabular}
\end{adjustbox}
\vspace{0.25em}
\footnotesize
\emph{Message:} field-aware learning is not a universal accuracy booster. It reduces field errors broadly, but improves ID/OOD only when aligned with path structure and appropriate optimization; otherwise it can over-regularize or trade task performance for field quality.
\end{table*}

\begin{table*}[t]
\centering
\caption{
\textbf{Continual learning with field preservation.}
Split CIFAR-100 is divided into 20 sequential tasks. Phase 0 diagnoses whether field drift correlates with forgetting; Phases 1--3 evaluate standalone FPR and its budget behavior; Phase 4 adds FPR as a field-retention regularizer to replay/distillation methods. Higher AA and BWT are better; lower FRS/JRS are better. In Phase 4, positive \(\Delta\)AA/\(\Delta\)BWT and negative \(\Delta\)FRS/\(\Delta\)JRS indicate improvement.
}
\label{tab:app_cl_fpr_all}
\scriptsize
\setlength{\tabcolsep}{4pt}
\renewcommand{\arraystretch}{1.05}
\begin{adjustbox}{width=\textwidth}
\begin{tabular}{lllcc}
\toprule
\textbf{Phase} & \textbf{Setting} & \textbf{Method / Test} & \textbf{Function metric} & \textbf{Field metric} \\
\midrule

\multirow{3}{*}{0}
& \multirow{3}{*}{Drift diagnosis}
& Trajectory drift vs. drop
& \(r=0.1748,\ p=0.0158\)
& \(\rho=0.1498,\ p=0.0391\) \\
& & Jacobian drift vs. drop
& \(r=0.0266,\ p=0.7159\)
& \(\rho=0.0070,\ p=0.9235\) \\
& & Parameter drift vs. drop
& \(r=-0.0853,\ p=0.2418\)
& \(\rho=-0.0924,\ p=0.2046\) \\

\midrule
\multirow{8}{*}{1}
& \multirow{8}{*}{Method comparison}
& Finetune
& AA \(0.0384\), BWT \(-0.6437\)
& FRS \(5.7340\), JRS \(0.1193\) \\
& & EWC
& AA \(0.0140\), BWT \(-0.0921\)
& FRS \(7.2487\), JRS \(0.1195\) \\
& & LwF
& AA \(0.0467\), BWT \(-0.6925\)
& FRS \(1.5442\), JRS \(0.1183\) \\
& & ER
& AA \(0.1160\), BWT \(-0.4955\)
& FRS \(0.7644\), JRS \(0.1610\) \\
& & DER++
& AA \(\mathbf{0.1346}\), BWT \(-0.5248\)
& FRS \(0.9209\), JRS \(0.1768\) \\
& & FPR-Traj
& AA \(0.0374\), BWT \(-0.6546\)
& FRS \(0.5568\), JRS \(\mathbf{0.0332}\) \\
& & FPR-Jac
& AA \(0.0345\), BWT \(-0.5707\)
& FRS \(1.2716\), JRS \(0.0536\) \\
& & FPR-Full
& AA \(0.0360\), BWT \(-0.6263\)
& FRS \(1.5758\), JRS \(0.0404\) \\

\midrule
\multirow{6}{*}{2}
& \multirow{6}{*}{FPR ablation}
& FPR-Full
& AA \(0.0353\), BWT \(-0.6360\)
& FRS \(\mathbf{0.4083}\), JRS \(\mathbf{0.0347}\) \\
& & FPR-TrajOnly
& AA \(0.0368\), BWT \(-0.6534\)
& FRS \(13.5703\), JRS \(0.1777\) \\
& & FPR-JacOnly
& AA \(0.0334\), BWT \(-0.5735\)
& FRS \(1.9565\), JRS \(0.0539\) \\
& & FPR-Early
& AA \(0.0289\), BWT \(-0.6182\)
& FRS \(2.0783\), JRS \(0.0743\) \\
& & FPR-Mid
& AA \(0.0382\), BWT \(-0.6582\)
& FRS \(3.4924\), JRS \(0.0780\) \\
& & FPR-Late
& AA \(0.0339\), BWT \(-0.5223\)
& FRS \(4.3175\), JRS \(0.1226\) \\

\midrule
\multirow{4}{*}{3}
& \multirow{4}{*}{Budget}
& 50 samples/task
& AA: ER \(0.0584\), DER++ \(0.0712\), FPR \(0.0367\)
& JRS: ER \(0.1715\), DER++ \(0.1675\), FPR \(\mathbf{0.0447}\) \\
& & 100 samples/task
& AA: ER \(0.0897\), DER++ \(0.1080\), FPR \(0.0375\)
& JRS: ER \(0.1552\), DER++ \(0.2443\), FPR \(\mathbf{0.0500}\) \\
& & 200 samples/task
& AA: ER \(0.1156\), DER++ \(0.1308\), FPR \(0.0321\)
& JRS: ER \(0.2074\), DER++ \(0.2062\), FPR \(\mathbf{0.0448}\) \\
& & 500 samples/task
& AA: ER \(0.1596\), DER++ \(0.1903\), FPR \(0.0361\)
& JRS: ER \(0.2391\), DER++ \(0.2132\), FPR \(\mathbf{0.0600}\) \\

\midrule
\multirow{6}{*}{4}
& \multirow{6}{*}{Hybrid FPR}
& DER++ \(\rightarrow\) DER++ + FPR-Full, 50
& \(\Delta\)AA \(-0.0021\), \(\Delta\)BWT \(-0.0218\)
& \(\Delta\)FRS \(-0.2579\), \(\Delta\)JRS \(-0.1279\) \\
& & DER++ \(\rightarrow\) DER++ + FPR-Full, 100
& \(\Delta\)AA \(+0.0008\), \(\Delta\)BWT \(-0.0429\)
& \(\Delta\)FRS \(-0.1135\), \(\Delta\)JRS \(-0.1176\) \\
& & DER++ \(\rightarrow\) DER++ + FPR-Full, 200
& \(\Delta\)AA \(\mathbf{+0.0111}\), \(\Delta\)BWT \(\mathbf{+0.0571}\)
& \(\Delta\)FRS \(\mathbf{-0.2661}\), \(\Delta\)JRS \(\mathbf{-0.0748}\) \\
& & ER \(\rightarrow\) ER + FPR-Late, 50
& \(\Delta\)AA \(-0.0267\), \(\Delta\)BWT \(-0.0581\)
& \(\Delta\)FRS \(+0.5128\), \(\Delta\)JRS \(+0.0012\) \\
& & ER \(\rightarrow\) ER + FPR-Late, 100
& \(\Delta\)AA \(-0.0470\), \(\Delta\)BWT \(-0.1027\)
& \(\Delta\)FRS \(+0.4941\), \(\Delta\)JRS \(+0.0028\) \\
& & ER \(\rightarrow\) ER + FPR-Late, 200
& \(\Delta\)AA \(-0.0461\), \(\Delta\)BWT \(-0.1415\)
& \(\Delta\)FRS \(+0.3111\), \(\Delta\)JRS \(-0.0207\) \\

\bottomrule
\end{tabular}
\end{adjustbox}
\vspace{0.25em}
\footnotesize
\emph{Message:} standalone FPR strongly preserves propagation-field metrics, especially JRS, but does not replace replay or distillation for function retention. As a regularizer, FPR is most effective when combined with a strong base method: DER++ + FPR-Full at 200 samples/task improves AA, BWT, FRS, and JRS simultaneously.
\end{table*}

%%%%%%%%%%%%%%%%%%%%%%%%%%%%%%%%%%%%%%%%%%%%%%%%%%%%%%%%%%%%%
%%%%%%%%%%%%%%%%%%%%%%%%%%%%%%%%%%%%%%%%%%%%%%%%%%%%%%%%%%%%%%%%%%%%%%%%%%%%%%%

\end{document}